\documentclass{article}

\usepackage{arxiv}

\usepackage{microtype}
\usepackage{graphicx}
\usepackage{caption}
\usepackage{subcaption}
\usepackage{booktabs} % for professional tables
\usepackage{hyperref}
\usepackage{amsmath}
\usepackage{amssymb}
\usepackage{mathtools}
\usepackage{amsthm}
\usepackage{mathabx}
\usepackage{bbm}
\usepackage{float}

\usepackage[capitalize,noabbrev]{cleveref}
\usepackage[textsize=tiny]{todonotes}
\DeclareMathOperator*{\argmin}{argmin}

\newcommand{\ep}{\mathbb{E}}

\newcommand\expe[1]{\mathop{\ep}_{#1}}
\newcommand\tin[1]{\text{\textit{\tiny{#1}}}}

\newcommand\bhf{\bar{\hat{f}}}
\newcommand\hfs{\hat{f_{\mathcal{S}}}}

\newcommand\hYsm{\hat{Y}_{\mathcal{S}_m}}

\newcommand\bhY{\bar{\hat{Y}}}
\newcommand\bhy{\bar{\hat{y}}}

\newcommand\Yhs{\hat{Y}_{\mathcal{S}}}
\newcommand\bx{\mathbf{x}}
\newcommand\cX{\mathcal{X}}

\DeclareMathAlphabet\mathbfcal{OMS}{cmsy}{b}{n}

\newcommand\bhYmm{\bhY_{\frac{m_1}{m_0}}}
\newcommand\hYmm{\hat{Y}_{\mathcal{S}_{\frac{m_1}{m_0}}}}
\newcommand\bhYmmp{\bhY_{\frac{m^p_1}{m^p_0}}}

\theoremstyle{plain}
\newtheorem{theorem}{Theorem}[section]

\theoremstyle{definition}
\newtheorem{definition}[theorem]{Definition}

\theoremstyle{remark}

\title{Shedding light on underrepresentation and Sampling Bias in machine learning}

\author{
  Sami Zhioua\\
  INRIA, LIX, École Polytechnique \\
  Palaiseau, Paris, France\\
  \texttt{zhioua@lix.polytechnique.fr} \\
   \And
  Rūta Binkytė \\
  INRIA, LIX, École Polytechnique \\
  Palaiseau, Paris, France\\
  \texttt{ruta.binkyte@inria.fr} \\
}

\begin{document}
\maketitle

\begin{abstract}

Accurately measuring discrimination is crucial to faithfully assessing fairness of trained machine learning (ML) models. Any bias in measuring discrimination leads to either amplification or underestimation of the existing disparity. %discrimination\todo[inline]{"disparity"?}. 
Several sources of bias exist and it is assumed that bias resulting from machine learning is born equally by different groups (e.g. females vs males, whites vs blacks, etc.). If, however, bias is born differently by different groups, it may exacerbate discrimination against specific sub-populations. % \todo[inline]{subpopulations? to avoid repetition}groups. 
Sampling bias, is inconsistently used in the literature to describe bias due to the sampling procedure. In this paper, we attempt to disambiguate this term by introducing clearly defined variants of sampling bias, namely, sample size bias (SSB) and underrepresentation bias (URB). We show also how discrimination can be decomposed into variance, bias, and noise. Finally, we challenge the commonly accepted mitigation approach that discrimination can be addressed by collecting more samples of the underrepresented group.
\end{abstract}

% keywords can be removed

\section{Introduction}
\label{sec:intro}

With the ubiquitous use of machine learning (ML) systems to inform decisions with critical impacts on human lifes (e.g. job hiring, college admission, security screening), fairness is emerging as an important requirement for the safe use of these technologies. A failure to guarantee fairness may create or amplify discrimination against individuals or specific sub-populations (e.g. minority groups). Such anomaly can initiate a vicious cycle that can be perpetuated and eventually resulting in severe consequences. 

Discrimination in ML decisions can originate from several types of bias as described in the literature. For instance, The Centre for Evidence-Based Medicine (CEBM) at the University of Oxford is maintaining a list of $62$ different sources of bias~\cite{oxfordCatalogueBias}. More related to ML, Mehrabi et al.~\cite{mehrabi2021survey} classify the sources of bias into three categories depending on when the bias is introduced in the automated decision loop. For instance, measurement bias~\cite{hellstrom2020bias,suresh2019framework} can be introduced at the data generation step and is a result of measuring a feature using a proxy variable instead of an ideal variable (e.g. using SAT score variable as a measure for the qualification feature). %For instance, typically in job hiring scenarios, qualification can be measured using SAT score variable which is clearly a non-reliable measure of qualification. 

Another, more common, category of bias occurs when the ML model is trained using a limited number of samples. This produces an inaccurate model and the inaccuracy will typically be born differently by different sub-populations which leads to a discrimination. Two famous examples of ML discrimination fall into this category of bias. The first is COMPAS software~\cite{COMPAS} used by several states in US to help predict whether a defendant will recidivate in the next two years if she is released. The software is found to be discriminatory against african-americans as the false positive rate (FPR) was higher for african-americans compared to other ethnicities, but the false negative rate (FNR) was lower~\cite{angwin2016machine}. The second example is related to face recognition technology (FRT). Buolamwini et al.~\cite{buolamwini2018gender} found that several commercial FRT software have a significantly lower accuracy for individuals belonging to a specific sub-population, namely, dark-skinned females. 

%, for which can When a model is learned using a  is available originates in how data is sampled, in particular, the limited number of samples that can affect the entire population or only specific sub-populations. 
This category of bias is inconsistently given various names in the literature (e.g. sampling bias, representation bias, data imbalance bias, etc.) and, to the best of our knowledge, is not formally defined. This paper is an attempt to disambiguate this category of bias by proposing definitions of two sources of bias, namely, sample size bias (SSB) and underrepresentation bias (URB). SSB is the bias that results from training an ML model using a training data with a limited number of samples and where all sub-populations are represented in the same proportions as the real population. %equally represented \todo[inline]{we need to mae a distinction here, as it now suggests that they are 50/50. It should be represented with proportions the same as in the population}. 
URB is the bias resulting from training an ML model using a training data with a disparity in the number of samples corresponding to each sub-population. 

Although the link between the limited number of samples used for training and the disparity in the accuracy of the obtained model may seem straightforward, the magnitude of such pattern has not been thoroughly studied in the ML fairness literature. Based on the proposed definitions of SSB and URB, the empirical part of the paper tries to illustrate how the magnitude of discrimination behaves as more extreme versions of bias are considered. Several metrics of discrimination are used, namely, difference in  $FPR$ (false positive rate), equal opportunity~\cite{hardt2016equality}, difference in $ZOL$ (zero-one loss), difference in $AUC$ (area under the curve), statistical disparity~\cite{dwork2012fairness}, and, for regression problems, difference in $MSE$ (mean squared error). For the latter, we use previous results in the literature~\cite{chen2018,domingos2000} to decompose the discrimination into noise, bias, and variance. 

The very definition of sampling bias suggests that it can be mitigated by simply using more data for training, in particular for the under represented groups. Obtaining more data is possible either through data augmentation (duplicating or creating synthetic samples) or resuming data collection. Unlike data augmentation, whose effect on discrimination has been the topic of a number of papers, in particular related to computer vision (e.g.~\cite{pastaltzidis2022data,yucer2020exploring, zhang2020towards,xu2020investigating,Wang2018BalancedDA,wang2020towards}), the impact of  of collecting more samples on discrimination has not been well studied in the literature%~\footnote{Detailed related work is provided in Appendix~\ref{sec:a_related}.}
. The last part of the paper studies the effect of collecting more samples on discrimination.

The key findings of this paper are the following:
\begin{itemize}
\item Discrimination defined in terms of cost/accuracy metrics that consider a trade-off between precision and recall (e.g. $AUC$ and $ZOL$) are more resilient to limited size or imbalanced training sets.
\item For extremely small or imbalanced training sets, the variance component of the bias is significant and can significantly alter the fairness conclusions.
\item In presence of underrepresented groups, collecting more data samples for the underrepresented group typically amplifies discrimination rather than reduces it.
\end{itemize}

% \begin{enumerate}
%     \item Disambiguate the inconsistently overloaded \todo[inline]{overloading?} ``sampling bias'' term by defining sample size bias (SSB) and underrepresentation bias (URB)
%     \item Illustrate empirically how the magnitude of discrimination behaves as more extreme versions of bias are considered
%     \item Identify threshold values for bias parameters beyond which discrimination is significantly affected.
% \end{enumerate}

% It is assumed that bias resulting from machine learning is born equally by different groups. If bias is born differently by different group, it may lead to discrimination against specific groups.

% Any inaccuracy/deviation/bias in computing discrimination can amplify or underestimate the discrimination. Hence bias have a direct impact on discrimination.

% To avoid confusion in the used terms, we dedicate one section to explicitly define the different terms.

% By expressing mathematically the different types of bias, we try to compare their magnitudes.

% In the presence of several types of bias, can we distinguish between them to dissect the observed bias in terms of the different types of bias.

% Test \cite{mitchell80}.

\section{Related Work}
\label{sec:related}

An exhaustive list of sources of bias can be found in the survey paper of Mehrabi et al.~\cite{mehrabi2021survey} where the authors categorized them into three categories: biases in the data (measurement, representation\footnote{The term representation bias in the context of computer vision denotes a different type of bias: the presence of potential ``shortcuts'' that a model can exploit to accurately predict the label without learning the underlying task~\cite{li2019repair}. A simple example is when the background environment of a picture can be used to recognize the object of interest.}, etc.), biases in the algorithm (algorithmic, evaluation, etc.), and biases introduced by users (historical, self-selection, etc.). This paper focuses on biases in the first category. Suresh et al.~\cite{suresh2019framework}, however, categorized the sources of bias into five categories depending on where in the machine learning pipeline a bias may be introduced. Although they provide a framework for the machine learning pipeline transformations, sources of bias that can impact the transformations are only described informally (Figure 2 in~\cite{suresh2019framework}). Similarly, Hellstrom et al.~\cite{hellstrom2020bias} propose a taxonomy of the sources of bias while categorizing them on the basis of the machine learning pipeline. The main limitation of all previous work on sources of bias is the absence of formal definitions of biases. Therefore, these papers did not include an analysis of the correlation between the magnitude of the bias and the extent of the discrimination.

Sampling bias is related to the known problems of (1) learning using a limited size training set and (2) learning using imbalanced data~\cite{he2009learning}. Chen et al.~\cite{chen2018} studied the effect of the training set sample size on discrimination. They used two discrimination metrics, namely, false positive rate (FPR) and false negative rate (FNR). They found that discrimination according to FNR is much more sensitive to sample size than FPR. To address the problem of imbalanced data, Yan et al.~\cite{yan2020fair} compared existing techniques for balancing data and found that while they achieve better prediction, they tend to exacerbate discrimination. Farrand et al.~\cite{farrand2020neither} focused on the impact of using imbalanced data on the accuracy and fairness of the obtained model while learning with privacy (differencial privacy (DP)) constraints. They found that data imbalance has little effect on discrimination (equal opportunity and statistical disparity) until the imbalance between sensitive groups becomes extreme (e.g. $99.9\%$ vs $0.1\%$). Interestingly, the impact is more important for DP-learned models than with non-DP models.

Machine learning loss/error has been first decomposed into bias and variance by Dietterich and Kong~\cite{dietterich1995}. The decomposition did not distinguish between variance and noise and considered noise as part of variance (variance is defined as the difference between loss and bias). They illustrated the decomposition for regression and for classification problems but only for individual examples. Domingos~\cite{domingos2000} distinguished between noise and variance and extended the decomposition to the expectation over all samples. He considered three loss functions, namely, squared loss, absolute loss, and zero-one loss. For the experimental analysis, Domingos focused on decision trees and k-nearest neighbor (KNN) learning algorithms and studied the effect of some parameters on the loss, namely, pruning parameters, level of the tree, number of rounds in boosting, and the k parameter in KNN. Chen et al.~\cite{chen2018} leveraged these previous results to decompose the discrimination between sensitive groups. They considered two types of loss functions, namely, zero-one loss for classification and squared loss for regression. They also studied the effect of increasing training data size on the discrimination. To this end, they assumed that the losses (population and group-specific) have an inverse power-low behavior asymtotically. This allows to predict the discrimination level if training data is augmented with additional samples.

Additional data collection for minority group or data augmentation is well known as a fairness remedy for imbalanced data, particularly in computer vision~\cite{pastaltzidis2022data,yucer2020exploring,zhang2020towards,xu2020investigating,Wang2018BalancedDA,wang2020towards, buolamwini2018gender}. Data augmentation can be split into two approaches. First is replicating or generating additional synthetic data points for the minority group that follows the same distribution~\cite{iosifidis2018dealing}. Second approach for data augmentation consists of selectively~\textit{balancing} the data with respect to the positive label and the sensitive group\cite{burns2018women}. In this study we explore the increase in data size, or the size of a sensitive minority group by adding additional samples from the same population distribution, which is closer to the first approach or additional data collection.

%\section{Preliminaries}
% \input{text-prelim}

\section{Preliminaries}

Let $\mathcal A$ be a supervised learning algorithm for learning an unknown function $f: \mathbfcal{X} \mapsto \mathcal{Y}$ where $\mathbfcal X$ is the input variables space and $\mathcal Y$ is the outcome space. Without loss of generality, the outcome random variable $Y$ is assumed to be binary ($\mathcal{Y} = \{0,1\}$, e.g. accepted/rejected). 
%Let $\mathcal D$ be a probability distribution over $\mathbfcal{X}$ used to draw random samples with probability $\mathcal{D}(\mathbf{x})$. 
Let $\mathcal{S} = \{(\mathbf{x}_i,y_i=f(\mathbf{x}_i))\}, i=1\ldots m$, be a training sample of size $m$. % drawn according to the distribution $\mathcal{D}$. 
Based on the data sample $\mathcal S$, algorithm $\mathcal A$ learns a function $\mathcal{A}(\mathcal{S}) = \hat{f}_{\mathcal{S}}^{\mathcal{A}}$. Let $\hat{Y}_{\mathcal{S}}^{\mathcal{A}}$ be the predicted outcome random variable such that $\hat{f}_{\mathcal{S}}^{\mathcal{A}}(\mathbf{x}_i) = \hat{y_i}$. When there is no ambiguity, we refer to $\hat{Y}_{\mathcal{S}}^{\mathcal{A}}$ and $\hat{f}_{\mathcal{S}}^{\mathcal{A}}$ simply as $\hat{Y}$ (or $\Yhs$) and $\hat{f}$ (or $\hat{f}_{\mathcal{S}}$).

Given the true value $y$ and the prediction $\hat{y}$, $L(y,\hat{y})$ represents the loss incurred by predicting $\hat{y}$ while the true outcome is $y$. A commonly used loss function for regression problems is the squared loss defined as $L^{SL}(\hat{y},y) = (\hat{y}-y)^2$. Other loss functions that will be considered in this paper are the absolute loss $L^{AL}(\hat{y},y) = |\hat{y}-y|$ and the zero-one loss $L^{ZO}(\hat{y},y) = 0$ if $\hat{y} = y$, and $1$ otherwise.

Based on a loss function, we define two special predictions, namely, the main prediction for a learning algorithm $\mathcal{A}$ and the optimal prediction. 

Given a learning algorithm $\mathcal{A}$ and a set of training samples $\mathfrak{S} = \{\mathcal{S}_1, \mathcal{S}_2, \ldots\}$, the main prediction random variable $\bhY^{\mathcal{A}}_{\mathfrak{S}}$ ($\bhy = \bhf_{\mathfrak{S}}^\mathcal{A}(\bx)$) represents the prediction that minimizes the loss across all training sets in $\mathfrak{S}$. That is, 
%$$\bhy^{\mathfrak{S}} = \argmin_{y'} \ep_{\mathfrak{S}} [L(\yhs,y)]$$ 
$$\bhf^{\mathfrak{S}}_{\mathcal{A}}(\bx) = \argmin_{f'} \ep_{\mathcal{S} \in \mathfrak{S}} [L(\hfs(\bx),f'(\bx)].$$ 
When there is no ambiguity, we refer to $\bhY^{\mathcal{A}}_{\mathfrak{S}}$ and $\bhf_{\mathfrak{S}}^{\mathcal{A}}(\bx)$ simply as $\bhY$ and $\bhf(\bx)$.
Typically, the main prediction corresponds to the average prediction 
%For the squared loss $L^{SL}$, the main prediction is the average prediction 
across all training sets in $\mathfrak{S}$. That is,
\begin{equation}\bhf(\bx) =  \displaystyle \expe{\mathcal{S} \in \mathfrak{S}} \hat{f}_{\mathcal{S}}(\mathbf{x})\footnote{We exceptionally use the expectation on a function, instead of a random variable.}.\end{equation}

The optimal prediction $Y^*$ ($y^* = f^{*}(\bx)$) is the prediction that minimizes the loss across all possible predictors. That is,
$$f^{*}(\bx) = \argmin_{f'} \ep[L(f(\bx),f'(\bx)].$$
%$$f^{*}(\bx) = \argmin_{f'} \ep_{Y} [L(f(\bx),f'(\bx)].$$
It is important to note that $f^{*}$ is independent of the learning algorithm $\mathcal{A}$.

Assume that the sensitive attribute $A$ is a binary variable with possible values $A=a_0$ and $A=a_1$, each representing a different group (e.g. male vs female, black vs white, etc.). Let $G_0$ and $G_1$ denote these groups. That is, $G_0 = \{\bx \in \cX | A=a_0\}$ and $G_1 = \{\bx \in \cX | A=a_1\}$. Discrimination between $G_0$ and $G_1$ can be defined in terms of the disparity in prediction accuracy. Let $C_a^{\bullet}(\hat{Y})$ denote the accuracy/cost of prediction $\hat{Y}$ for group $A=a$. For classification problems, we consider four metrics, namely, false positive rate ($FPR$), false negative rate ($FNR$), true positive rate ($TPR$), and zero one loss ($ZOL$). For regression problems, we consider mean square error ($MSE$). These metrics are defined as follows:
\begin{itemize}
    \item[$\circ$] $C^{\tin{FPR}}_a(\hat{Y}) = \ep[\hat{Y} | Y=0, A=a]$
    \item[$\circ$] $C^{\tin{FNR}}_a(\hat{Y}) = \ep[1-\hat{Y} | Y=1, A=a]$
    \item[$\circ$] $C^{\tin{TPR}}_a(\hat{Y}) = \ep[\hat{Y} | Y=1, A=a]$
    \item[$\circ$] $C^{\tin{ZOL}}_a(\hat{Y}) = \ep[\mathbbm{1}[\hat{Y}\neq Y]| A=a]$
    \item[$\circ$] $C^{\tin{MSE}}_a(\hat{Y}) = \ep[(\hat{Y} - Y)^2| A=a]$
\end{itemize}

Discrimination $Disc^{\bullet}$ can be defined as the difference in $C_a^{\bullet}$ between the two sensitive groups. For instance $Disc^{\tin{FPR}}(\hat{Y}) = C_{a_1}^{\tin{FPR}}(\hat{Y}) - C_{a_0}^{\tin{FPR}}(\hat{Y})$. Notice that $Disc^{\tin{TPR}}(\hat{Y})$ corresponds to discrimination according to equal opportunity~\cite{hardt2016equality} and that $Disc^{\tin{TPR}}(\hat{Y}) = - Disc^{\tin{FNR}}(\hat{Y})$ as $TPR = 1 - FNR$. In the rest of the paper, we use $Disc^{\tin{TPR}}(\hat{Y})$ and $Disc^{\tin{EO}}(\hat{Y})$ interchangeably. In addition, for reference, we use $Disc^{\tin{SD}}(\hat{Y}) = \ep[\hat{Y}|A=a_1] - \ep[\hat{Y}|A=a_0]$ to denote statistical disparity~\cite{dwork2012fairness}.

% In particular, for regression problems (the outcome $Y$ is a continuous value such as a score), discrimination can be defined as the disparity in the mean squared error (MSE). That is, 
% \begin{equation} Disc^{MSE}(\hfs) = \overline{L^{SL}_{a_1}}(\hfs) - \overline{L^{SL}_{a_1}}(\hfs)\label{eq:disc0}\end{equation}
% where:
% \begin{itemize}
%     \item[$\circ$] $\overline{L^{SL}_{a_1}}(\hfs) = \ep_{\bx \in \mathcal{X}}[(\hfs(\bx) - f(\bx))^2]$
% \end{itemize}

\section{Sample Size and Underrepresentation Biases}

%\subsection{Sample Size Bias}

Typically, the size of the data used to train an ML model has a significant impact on the accuracy of the obtained model. However, it is generally assumed that the loss in accuracy is equally born by the different segments of the data. As it is not usually the case, we define sample size bias (SSB) as the bias resulting from training a model with a given data size.
%Training an ML model using a training set of a limited size has 

%Sample size bias is caused by the limited size of the training data. 

Let $\mathfrak{S_m} = \{\mathcal{S}_1, \mathcal{S}_2, \ldots\}$ be the set of samples of size $m$, and let $\hat{f}_{\mathcal{S}_1}, \hat{f}_{\mathcal{S}_2}, \ldots$ be the models produced by applying the learning algorithm $\mathcal A$ on each sample ($\mathcal{A}(\mathcal{S}_1)=\hat{f}_{\mathcal{S}_1}$, etc.). Let $\bhY^{\mathcal{A}}_{\mathfrak{S_m}}$ ($\bhy_m = \bhf_{\mathfrak{S}_m}^{\mathcal{A}}(\bx)$) be the main prediction obtained using the set of training sets $\mathfrak{S_m}$. That is,
\begin{equation}\bhf_{\mathfrak{S}_m}^{\mathcal{A}}(\bx) = \argmin_{f'} \ep_{\mathcal{S} \in \mathfrak{S}_m} [L(\hfs(\bx),f'(\bx)].
\end{equation}
When there is no ambiguity, we refer to $\bhY^{\mathcal{A}}_{\mathfrak{S_m}}$ and $\bhf_{\mathfrak{S}_m}^{\mathcal{A}}$ simply as $\bhY_m$ and $\bhf_m$.

\begin{definition}
Given a positive number $m > 0$ representing the training set size, sample size bias is the difference in discrimination due to the training set size:
\begin{equation}
    SSB^{\bullet}(\mathcal{A},m) = Disc^{\bullet}(\bhY_m) - Disc^{\bullet}(\bhY_{\infty}) \label{eq:ssb0}
\end{equation}
where $Disc^{\bullet}(\bhY_{\infty}) = \displaystyle \lim_{m\to \infty} Disc^{\bullet}(\bhY_m) $ and $\bullet$ is a placeholder for the accuracy/cost metric ($FPR, FNR, EO, ZOL,$ or $MSE$ for regression problems). As a metric that combines both specificity ($FPR$) and sensitivity ($TPR$), we use also $AUC$ (area under the curve)\footnote{Other metrics combining specificity and sensitivity include $F_1$ score and balanced accuracy ($BA$)}. For reference, we consider also statistical disparity that we denote as $Disc^{SD}$ (See Appendix~\ref{sec:sd}).  % is the prediction model function obtained by using an infinitely large training set.
\end{definition}

As $SSB$ is defined in terms of an infinite size training set ($\bhY_{\infty}$), we consider an alternative definition in terms of $M$, the size of the largest training set available:
\begin{equation}
    SSB_M^{\bullet}(\mathcal{A},m) = Disc^{\bullet}(\bhY_m) - Disc^{\bullet}(\bhY_M) \label{eq:ssb1}
\end{equation}

Another variant of $SSB$ can be defined based on a specific training set $\mathcal{S}_m$ of size $m$ as follows:
\begin{equation}
    SSB_M^{\bullet}(\mathcal{A},\mathcal{S}_m) = Disc^{\bullet}(\hYsm) - Disc^{\bullet}(\bhY_M) \label{eq:ssb2}
\end{equation}
%where $\mathcal{S}_M$ is a specific training set of size $M$. 

% Let $\mathfrak{S_m} = \{\mathcal{S}_1, \mathcal{S}_2, \ldots\}$ be the set of samples of size $m$, $\hat{f}_{\mathcal{S}_1}, \hat{f}_{\mathcal{S}_2}, \ldots$ be the models produced by applying the learning algorithm $\mathcal A$ on each sample ($\mathcal{A}(\mathcal{S}_1)=\hat{f}_{\mathcal{S}_1}$, etc.), and $\overline{\hat{f}}$ be the model obtained by averaging over all models:

%\subsection{Underrepresentation Bias}

When sampling a training set from a population, it is generally assumed that the generated sample is balanced. Data is balanced if all classes are proportionally represented and is imbalanced if it suffers from severe class distribution skews~\cite{he2009learning}. For instance, if one class label is overrepresented at the expense of another underrepresented class label. If data is imbalanced in the sensitive groups (e.g. male vs female, blacks vs whites, etc.), it can have significant impact on the disparity of accuracies and consequently on discrimination between sensitive groups. We define underrepresentation bias (URB) as the bias resulting from a disparity in representation between the sensitive groups. 

Let $\mathfrak{S^{\frac{m_1}{m_0}}_m}$ be the set of samples of size $m$ with $m_0$ and $m_1$ items from $G_0$ and $G_1$ respectively. That is, for $\mathcal{S} \in \mathfrak{S^{\frac{m_1}{m_0}}_m}$, $|\{\bx \in \mathcal{S} | A=a_0\}| = m_0$, $|\{\bx \in \mathcal{S} | A=a_1\}| = m_1$, and $m_0 + m_1 = m = |\mathcal{S}|$. We use the simpler notation $\bhY_{\frac{m_1}{m_0}}$ to refer to $\bhY_{\mathfrak{S^{\frac{m_1}{m_0}}_m}}^{\mathcal{A}}$.

\begin{definition}
\label{def:urb}
Given, $m, m_0, m_1 > 0$ such that $m_0 + m_1 = m$, underrepresentation bias is the difference in discrimination due to the disparity in sample sizes compared to the population ratio:

\begin{equation}
 URB^{\bullet}(\mathcal{A},m_0,m_1) = Disc^{\bullet}(\bhYmm) - Disc^{\bullet}(\bhYmmp)   
\end{equation}
where $Disc^{\bullet}(\bhY_{{m^p_1}/{m^p_0}})$ is the discrimination of the prediction based on a model trained using only samples from $\mathfrak{S^{{m^p_1}/{m^p_0}}_m}$, and the ratio $\frac{m^p_1}{m^p_0}$ is the same as the ratio in the population ($\frac{m^p_1}{m^p_0} \approx \frac{|G_1|}{|G_0|}$).
\end{definition}

Similar to $SSB_M^{\bullet}(\mathcal{A},\mathcal{S}_m)$ (Equation~\ref{eq:ssb2}), a variant of $URB$ can be defined based on a specific training set $\mathcal{S}_{\frac{m_1}{m_0}} \in \mathfrak{S^{\frac{m_1}{m_0}}_m}$ as follows:
\begin{equation}
    URB^{\bullet}(\mathcal{A},\mathcal{S}_{\frac{m_1}{m_0}}) = Disc^{\bullet}(\hYmm) - Disc^{\bullet}(\bhYmmp) \label{eq:urb2}
\end{equation}

\section{Loss and Discrimination Decomposition}

\label{sec:decomposition}

Domingos~\cite{domingos2000} showed that if a learning algorithm $\mathcal A$ learns a function $\mathcal{A}(\mathcal{S}) = \hat{f}_{\mathcal{S}}$ based on a training set $\mathcal{S} \in \mathfrak{S}$, then the expected loss between the prediction $\hfs(\bx)$ and the true value $f(\bx)$ can be decomposed into noise, bias, and variance. In particular, for squared loss,
\begin{equation}L^{SL}(\hfs(\bx),f(\bx)) = N^{SL}(\bx) + B^{SL}(\bx) + V^{SL}(\bx)\label{eq:decomposition0}\end{equation}
where 
\begin{itemize}
    \item[$\circ$] $N^{SL}(\bx) = L^{SL}(f^{*}(\bx),f(\bx))$
    \item[$\circ$] $B^{SL}(\bx) = L^{SL}(\bhf(\bx),f^{*}(\bx))$
    \item[$\circ$] $V^{SL}(\bx) = \ep_{\mathcal{S} \in \mathfrak{S}}[L^{SL}(\hfs(\bx),\bhf(\bx))]$
\end{itemize}
The loss decomposition can be illustrated as follows:

\begin{tikzpicture}
\matrix[row sep=-0.2cm,column sep=0.2cm] {%
    % First row:
    \node (p1) {$f(\bx)$};   &&&& \node (p2) {$f^{*}(\bx)$}; &&&& \node (p3) {$\bhf(\bx)$}; &&&& \node (p4) {$\hfs(\bx)$}; \\
    && \node (p5) {$Noise$};   &&&& \node (p6) {$Bias$}; &&&& \node (p7) {$Variance$}; &&; \\
};
\draw   (p1) edge [<->,>=stealth,shorten <=2pt, shorten >=2pt, thick] (p2);
\draw   (p2) edge [<->,>=stealth,shorten <=2pt, shorten >=2pt, thick] (p3);
\draw   (p3) edge [<->,>=stealth,shorten <=2pt, shorten >=2pt, thick] (p4);       
\end{tikzpicture}

For Zero-One loss ($L^{ZO}$), Equation~\ref{eq:decomposition0} holds also but with coefficients different than $1$ for the noise and variance terms. However, it does not hold for the absolute loss ($L^{AL}$)\footnote{Alternatively, upper and lower bounds are possible.}~\cite{domingos2000}

% \begin{figure}[h]  
% \centering 

\subsection{Decomposing Discrimination}

Chen et al.~\cite{chen2018} showed that the accuracy/cost metric $C^{\bullet}_a(\Yhs)$ as well as the discrimination $Disc^{\bullet}(\Yhs)$ can be decomposed into noise, bias, and variance components. In particular, for MSE, 

\begin{equation}
    C^{\tin{MSE}}_a(\Yhs) = \overline{N}_{a}^{\tin{SL}}(\Yhs) + \overline{B}_{a}^{\tin{SL}}(\Yhs) + \overline{V}_{a}^{\tin{SL}}(\Yhs) \label{eq:LDecomp0}
\end{equation}
where:
\begin{itemize}
    \item[$\circ$] $\overline{N}_{a}^{\tin{SL}}(\Yhs) = \ep_{\bx \in \mathcal{X}} [N^{SL}(\bx)|A=a]$ 
    \item[$\circ$] $\overline{B}_{a}^{\tin{SL}}(\Yhs) = \ep_{\bx \in \mathcal{X}} [B^{SL}(\bx)|A=a]$ 
    \item[$\circ$] $\overline{V}_{a}^{\tin{SL}}(\Yhs) = \ep_{\bx \in \mathcal{X}} [(1-2\times B^{SL}(\bx))\times V^{SL}(\bx)|A=a]$    
\end{itemize}
The last term ($V_a(\Yhs)$) is called \textit{net variance}~\cite{domingos2000}. Consequently,
\begin{equation}
    Disc^{\tin{MSE}}(\Yhs) = (\overline{N}_{a_1}^{\tin{SL}}(\Yhs) - \overline{N}_{a_0}^{\tin{SL}}(\Yhs)) + (\overline{B}_{a_1}^{\tin{SL}}(\Yhs) - \overline{B}_{a_0}^{\tin{SL}}(\Yhs)) + (\overline{V}_{a_1}^{\tin{SL}}(\Yhs) - \overline{V}_{a_0}^{\tin{SL}}(\Yhs)) \label{eq:discDecomp0}
\end{equation}

 %Other notions of fairness can be defined using different accuracy cost functions such as $FPR$, $FNR$, and Zero-One loss ($ZOL$). 
 The decomposition of Equation~\ref{eq:discDecomp0} will also hold for $Disc^{\tin{FPR}}(\Yhs)$, $Disc^{\tin{EO}}(\Yhs)$, and $Disc^{\tin{ZOL}}(\Yhs)$ but with coefficients different than $1$ for the noise and variance terms~\cite{chen2018}.

\subsection{Decomposing $SSB$ and $URB$}
The variant $SSB_M^{\bullet}(\mathcal{A},\mathcal{S}_m)$ (Eq.~\ref{eq:ssb2}) of sample size bias has the advantage that it can be decomposed into bias and variance. The decomposition for the $MSE$ metric is as follows.
\begin{theorem}
\label{th:ssb_decomp}
$SSB_M^{\tin{MSE}}(\mathcal{A},\mathcal{S}_m)$ can be decomposed into bias and variance components as follows:
\begin{align}
SSB_M^{\tin{MSE}}(\mathcal{A},\mathcal{S}_m) & = \overline{B}^{SL}_{a_1}(\hYsm) - \overline{B}^{SL}_{a_1}(\bhY_M) - (\overline{B}^{SL}_{a_0}(\hYsm) - \overline{B}^{SL}_{a_0}(\bhY_M)) \nonumber \\
& + \overline{V}^{SL}_{a_1}(\hYsm) - \overline{V}^{SL}_{a_1}(\bhY_M) - (\overline{V}^{SL}_{a_0}(\hYsm) - \overline{V}^{SL}_{a_0}(\bhY_M)) \nonumber   
\end{align}
\end{theorem}
\begin{proof}
The proof follows from Equation~\ref{eq:LDecomp0} and from assuming that the optimal predictor $Y^{*}$ coincides with the true value $Y$ and hence noise is $0$ \footnote{We follow previous work (Domingos~\cite{domingos2000} and Kohavi and Wolpert~\cite{kohavi1996bias}) in assuming a zero noise.}. 
\end{proof}

$URB^{\bullet}(\mathcal{A},\mathcal{S}_{\frac{m_1}{m_0}})$ (Equation~\ref{eq:urb2}) can also be decomposed into bias and variance components. The decomposition for the $MSE$ metric is as follows.
\begin{theorem}
\label{th:urb_decomp}
\begin{align}
    URB^{\tin{MSE}}(\mathcal{A},\mathcal{S}_{\frac{m_1}{m_0}}) & = \overline{B}^{SL}_{a_1}(\hYmm) - \overline{B}^{SL}_{a_1}(\bhYmmp) - (\overline{B}^{SL}_{a_0}(\hYmm) - \overline{B}^{SL}_{a_0}(\bhYmmp)) \nonumber \\
& + \overline{V}^{SL}_{a_1}(\hYmm) - \overline{V}^{SL}_{a_1}(\bhYmmp) - (\overline{V}^{SL}_{a_0}(\hYmm) - \overline{V}^{SL}_{a_0}(\bhYmmp)) \nonumber 
\end{align}
\end{theorem}
\begin{proof}
The same as Theorem~\ref{th:ssb_decomp}.
\end{proof}

%$ = Disc^{\bullet}(\hYmm) - Disc^{\bullet}(\bhYmmp)$ \label{eq:urb2}

\section{Experimental Analysis}
% \begin{figure*}[t!]
%     \centering
%     \begin{subfigure}[t]{1\textwidth}
%         \centering
%         \includegraphics[height=2.5in]{figures/Sizes_adult_plot.pdf}
%         \caption{Adult}
%     \end{subfigure}%
%     ~ \\
%     \begin{subfigure}[t]{1\textwidth}
%         \centering
%         \includegraphics[height=2.5in]{figures/Sizes_compas_plot.pdf}
%         \caption{COMPAS}
%     \end{subfigure}
%     \caption{Sample size bias ($SSB$)}
%     \label{fig:ssb}
% \end{figure*}

\begin{figure*}[t!]
    \centering
    \includegraphics[height=1.9in, width=7.9in]{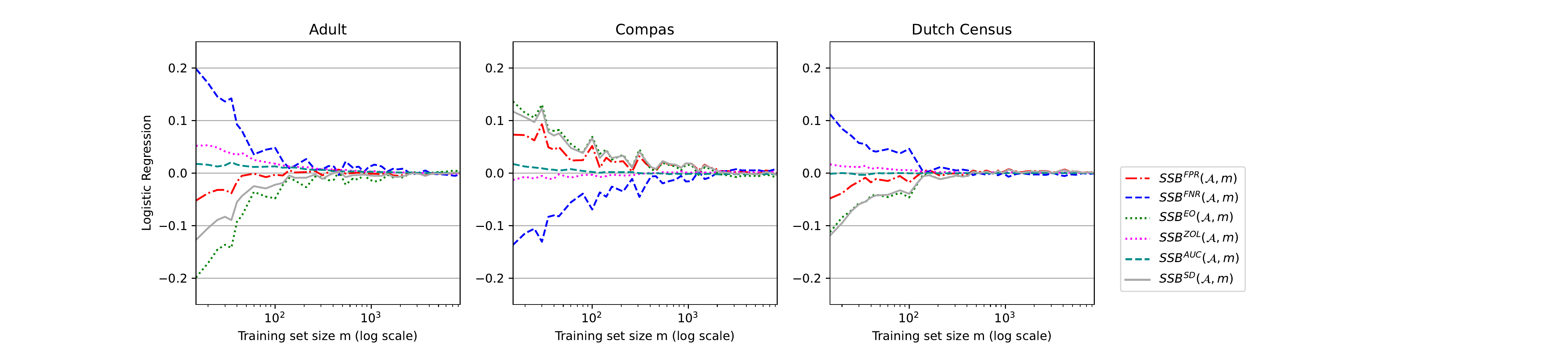}
    \caption{Magnitude of sample size bias (SSB) for increasing size of the training data.}
    \label{fig:ssb_lg}
\end{figure*}

The objective of the experimental analysis is to observe the magnitude of both types of biases, namely, sample size bias $SSB$ and underrepresenation bias $URB$ as we change the parameters of data sampling. For $SSB$, we train the predictor model using training sets of increasing sizes. For $URB$, we play rather on the proportions of sensitive groups in the training set. Three benchmark datasets are used, Adult~\cite{kohavi1996scaling}, Compas~\cite{angwin2016machine}, and Dutch Census~\cite{nordholt2004dutch}\footnote{We use the same dataset versions and learning algorithms parameters as IBM AIF360~\cite{ibm360}}. 

% The Adult dataset\footnote{https://archive.ics.uci.edu/ml/datasets/adult.} consists of $32,561$ samples and the goal is to predict the income of individuals based on several personal attributes such as age, race, sex, marital status, education, and employment. In this work, only $7$ variables are used for structural learning namely: age, sex, education level, marital status, work-class and number of working hours per week. The income of an individual can take two values namely: $\leq 50K$ (negative decision) or $>50K$ (positive decision). Age and number of working hours per week are continuous while the remaining variables are discrete. The COMPAS dataset includes data from Broward County, Florida, initially compiled by ProPublica~\cite{angwin2016machine} and the goal is to predict the two-year violent recidivism. That is, whether a convicted individual would commit a violent crime in the following two years ($1$) or not ($0$). Only black and white defendants who were assigned \textit{Compas} risk scores within $30$ days of their arrest are kept for analysis~\cite{angwin2016machine} leading to $5915$ individuals in total. We consider race as sensitive feature. The \textit{Dutch Census} dataset consists of $60,420$ tuples where the sensitive attribute is the sex of an individual and the outcome is her occupation (job). Four additional features are defined in the dataset, namely, 
% age, economic status, education, marital status and occupation. 

\subsection{Magnitude of sample size bias ($SSB$)}\label{sec:exp_ssb}
To observe how sample size bias behaves as the training set size changes, we use the following process. We use a sequence of sample sizes ranging from $10$ until a given portion of the full dataset size. For example, for COMPAS, we consider sample sizes ranging from $10$ to $2000$. For each sample size value $m$, we repeat the sampling several times ($30$ by default) so that we obtain $30$ samples of each size $m$. Then, we train a different model using each one of the samples so that we obtain $30$ models for each size $m$. We finally compute the dicrimination using each model and the returned value is the average discrimination across all models. This procedure gives a sequence of discrimination values indexed by the size. We consider five cost/accuracy metrics, namley, $FPR$ (false positive rate), $FNR$ (false negative rate), $EO$ (equal opportunity), $ZOL$ (zero one loss), and $SD$ (statistical disparity). We use five classifiers, namely, logistic regression, decision tree, random forest, nearst neighbor, and support vector machine (SVM). 
\begin{figure*}[t!]
    \centering
    \includegraphics[height=1.6in, width=7.7in]{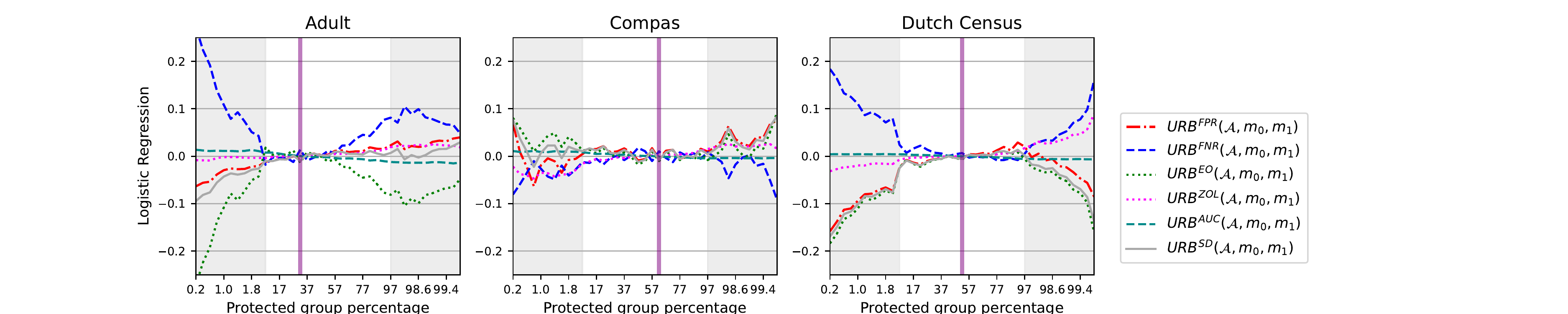}
    \caption{Underrepresentation Bias (URB) for different ratios of sensitive groups. The training set size is fixed ($1000$). The horizontal bar represents the same ratio as the population. The shaded sections indicate a focus on the extreme proportions (less than $2\%$ and more than $98\%$).}
    \label{fig:urb_lg}
\end{figure*}

Figure~\ref{fig:ssb_lg} shows the magnitude of SSB according to each metric and for each benchmark dataset and using logistic regression. Notice that $SSB^{EO}$ and $SSB^{FNR}$ are symmetric because, as mentioned above, $FNR = 1 - TPR$ and hence $SSB^{EO} = - SSB^{FNR}$. Most of the plots exhibit an expected behavior of SSB. That is, the bias is significant when the models are trained using a limited size training set. The bias disappears gradually as the training set size increases. $SSB$ behaves the same way for the other classifiers (Figure~\ref{fig:all_ssb} in Appendix~\ref{sec:a_ssb_urb}). More importantly, $SSB$ results show that cost/accuracy metrics that combine specificity and sensitivity ($AUC$ and $ZOL$) are less sensitive to the training set size than the remaining metrics ($FPR$ and $EO$). A possible explanation is that for small training sets, it is more likely that a majority of the samples have the same outcome (positive or negative) which can boost precision on the expense of recall or the opposite. $AUC$ and $ZOL$ are not subject to such skewness since they consider the trade-off between precision and recall.

\subsection{Magnitude of underrepresentation bias ($URB$)}\label{sec:exp_urb}

The aim for underrepresentation bias experiment is to observe the magnitude of $URB$ while the ratio of the sensitive groups in the training set is changing. We consider different values of the splitting $\frac{m_1}{m_0}$ (see Definition~\ref{def:urb}) (e.g. $0.1$ vs $0.9$, $0.2$ vs $0.8$, etc.). However, as $URB$ is more significant for extreme disparities, we focus more on extreme splitting values (e.g. $0.001$ vs $0.99$, $0.002$ vs $0.98$, etc.). A similar behavior has been observed previously by Farrand et al.~\cite{farrand2020neither}. Assuming a fixed sample size (e.g. $1000$), for each splitting value, we sample the data so that the proportions of sensitive groups (e.g. male vs female) match the splitting value. Similarly to the $SSB$ experiment, we repeat the sampling several times ($30$ by default) for the same splitting value. Then, we train a different model using each one of the samples so that we obtain $30$ models for each splitting value $\frac{m_1}{m_0}$. The discriminations obtained using the different models are then averaged across all models. We finally obtain a sequence of discrimination values indexed by the splitting value. Figure~\ref{fig:urb_lg} shows how URB changes as the proportion of the sensitive group increases for the same three datasets and for using logistic regression as learning algorithm. The purple vertical bar indicates the percentage of the sensitive group in the entire dataset (population). For instance, for adult dataset, the percentage of females is $31\%$. The shaded parts in the background of Figure~\ref{fig:urb_lg}'s plots indicate that we are ``zooming'' on the extreme values (the plots are using different steps for the shaded and unshaded parts\footnote{The step is very small below $2\%$ and above $98\%$.}). Almost all plots exhibit the same pattern for $URB$, that is, the further the proportions of sensitive groups are from the population proportions reference (vertical bar), the higher is the bias. The same expected behavior for $URB$ is obtained when using the other classifiers (Figure~\ref{fig:all_URB} in Appendix~\ref{sec:a_ssb_urb}). The resilience of $AUC$ and $ZOL$ metrics to extreme training set sizes holds also for imbalanced training sets. Notice that $URB^{AUC}$ and $URB^{ZOL}$ remain stable even for extremely imbalanced training sets.  

\begin{figure*}[t!]
    \centering
    \includegraphics[height=2.2in, width=6in]{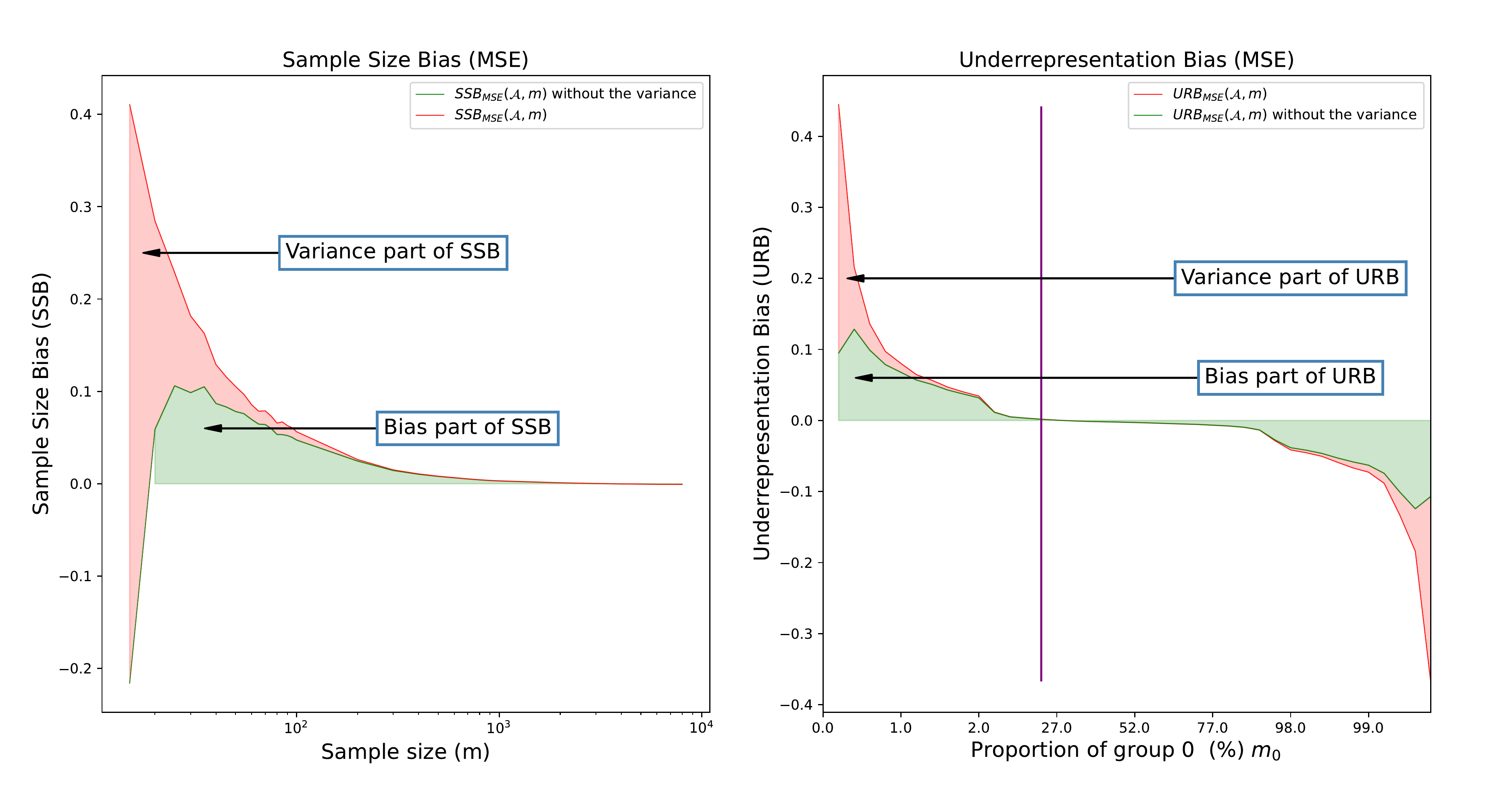}
    \caption{Decomposing $SSB^{MSE}$ (left plot) and $URB^{MSE}$ (right plot). The models are trained using linear regression. The benchmark dataset is Law School~\cite{wightman1998lsac}.}
    \label{fig:decomp_SSB_URB}
\end{figure*}

% \begin{figure*}[t!]
%     \centering
%     \includegraphics[height=1.4in, width=6in]{figures/contrSizes.pdf}
%     \caption{Sensitive feature importance while the training set size is increased.}
%     \label{fig:contrSizes}
% \end{figure*}

% \begin{figure*}[t!]
%     \centering
%     \includegraphics[height=1.4in, width=6in]{figures/contrRatios.pdf}
%     \caption{Sensitive feature importance while the ratio between sensitive groups is changing.}
%     \label{fig:contrRatios}
% \end{figure*}

% \begin{figure*}[t!]
%     \centering
%     \includegraphics[height=2.5in, width=5in]{figures/dutchSSB.pdf}
%     \caption{Discrimination and sample size bias (SSB) for the Dutch census dataset.}
%     \label{fig:dutchssb}
% \end{figure*}

% \begin{figure*}[t!]
%     \centering
%     \includegraphics[height=2.5in, width=5in]{figures/dutchURB.pdf}
%     \caption{Discrimination and underrepresentation bias (URB) for the dutch census dataset.}
%     \label{fig:dutchurb}
% \end{figure*}

% \begin{figure*}[t!]
%     \centering
%     \includegraphics[height=2.5in, width=5in]{figures/syntheticSSB.pdf}
%     \caption{Discrimination and sample size bias (SSB) for the synthetic dataset.}
%     \label{fig:syntheticssb}
% \end{figure*}

% \begin{figure*}[t!]
%     \centering
%     \includegraphics[height=2.5in, width=5in]{figures/syntheticURB.pdf}
%     \caption{Discrimination and underrepresentation bias (URB) for the synthetic dataset.}
%     \label{fig:syntheticurb}
% \end{figure*}

\subsection{Bias Decomposition}

Section~\ref{sec:decomposition} shows that loss and discrimination can be decomposed into variance, bias, and noise. In particular, assuming that the optimal prediction ($Y^{*}$) coincides with the correct outcome ($Y$),  Theorems~\ref{th:ssb_decomp} and~\ref{th:urb_decomp} illustrate how $SSB_M^{\tin{MSE}}(\mathcal{A},\mathcal{S}_m)$ and $URB^{MSE}(\mathcal{A},\mathcal{S}_{\frac{m_1}{m_0}})$ can be decomposed into variance and bias components. To illustrate the decomposition empirically, we use the Law School benchmark dataset~\cite{wightman1998lsac} which tracked some twenty-seven thousand law students through law school and graduation and where the sensitive attribute is gender and the outcome is the first year GPA. We use the scikit-learn linear regression algorithm to train different models using different size training sets. For $SSB$, the training size $m$ ranges from $10$ to $10,000$. For $URB$, the training set size ($m$) is fixed at $1000$, but the proportion of the protected group (female) is ranging from $0.1\%$ to $99.9\%$. For each training set size, the training and testing is repeated $30$ times. Figure~\ref{fig:decomp_SSB_URB} shows how $SSB$ and $URB$ are decomposed into variance and bias. For $SSB$, the variance component is so significant when the training set is extremely small (less than $20$) that it reverses the direction of the bias (in favor of females instead of against female). For $URB$, the variance is also significant when one of the groups is extremely underrepresented, but not to the point of reversing the direction of the bias. The main conclusion out of this emprical result is that for very small or very imbalanced training sets, $SSB$ and $URB$ variance can be so important that it can lead to unreliable conclusions about discrimination.

\subsection{Effect of collecting more samples on discrimination}

\label{sec:exp_threshold}

\begin{figure*}[t!]
    \centering
    \includegraphics[height=1.6in, width=7.2in]{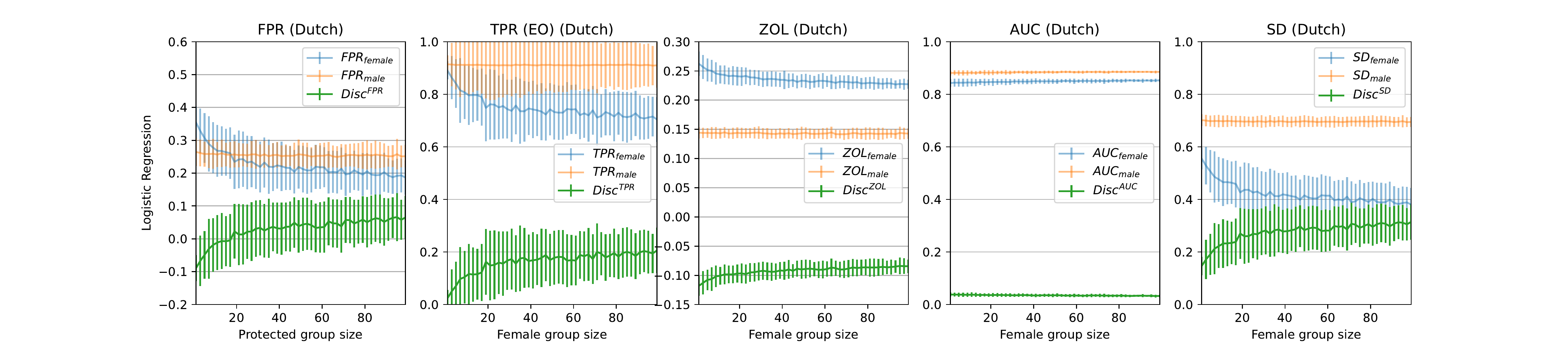}
    \caption{Discrimination while augmenting the training set with female group samples randomly. The male group size is fixed at $100$. Dataset is Dutch Census and training algorithm is logistic regression.}
    \label{fig:dutchThreshold}
\end{figure*}
The natural approach to address sampling bias is to use more data for training, in particular for the under-represented groups. Obtaining more data is possible either through data augmentation or data collection. Data augmentation is the process of using the available data to generate more samples. In turn, this can be done in two ways: oversampling or creating fake samples. Oversampling consists in duplicating existing samples to balance the data. A simple variant is to randomly duplicate samples from the under represented group. Creating fake samples, on the other hand, is typically done using SMOTE~\cite{chawla2002smote}. SMOTE creates synthetic samples based on the k-nearest neighbors of every sample of the under represented group. Both techniques of data augmentation try to balance data by adding artificially generated samples. While this artificial manipulation may reduce discrimination between sensitive groups, it can lead to models which are not faithful to reality. When it is possible, collecting more data is more natural and reflects better reality. The approach is simple: if a sensitive group is under represented, collect more samples of that group. Unlike data augmentation, whose effect on discrimination has been the topic of a number of papers, in particular related to computer vision (e.g.~\cite{pastaltzidis2022data,yucer2020exploring, zhang2020towards,xu2020investigating,Wang2018BalancedDA,wang2020towards}), the impact of  of collecting more samples on discrimination has not been well studied in the literature. 

\begin{figure*}[t!]
    \centering
    \includegraphics[height=1.6in, width=7in]{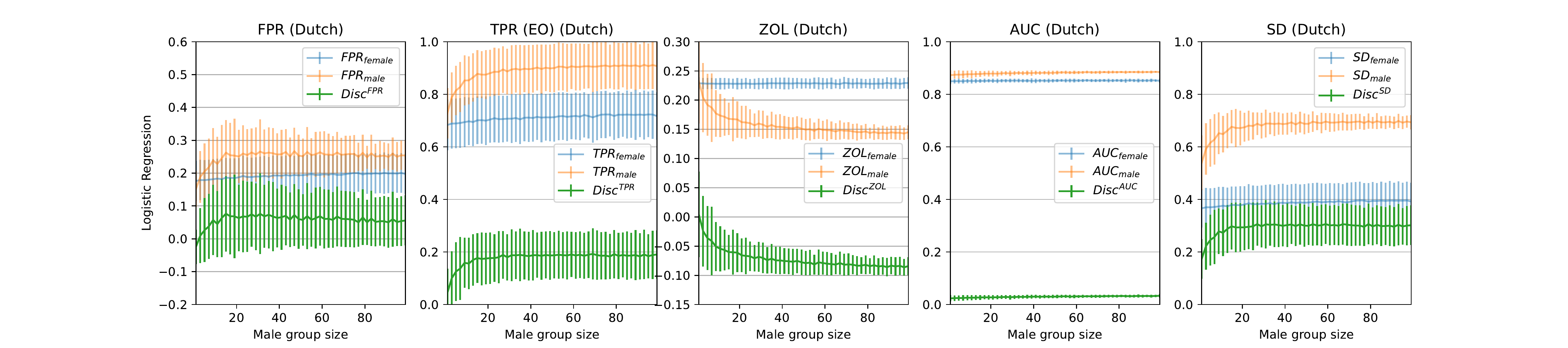}
    \caption{Discrimination while augmenting the training set with male group samples randomly. The female group size is fixed at $100$. Dataset is Dutch Census and training algorithm is logistic regression.}
    \label{fig:dutchThreshold_inv}
\end{figure*}

In the following, we devise simple experiments to observe the effect of populating the data with more samples collected from the same population as the existing data. Using the same benchmark datasets, the aim is to train models based on an increasing number of under represented group samples while keeping the privileged group portion unchanged. For the particular case of Dutch Census dataset, we train models using a set composed of a fixed $100$ privileged group (male) samples and an increasing number of protected group (female) samples starting from $2$ until $100$ (perfect balance between groups). Similarly to the SSB and URB experiments, it turns out that the magnitude of discrimination is manifested more with extreme values of protected groups sizes (typically less than $100$) which explains the specific sample sizes considered. Figure~\ref{fig:dutchThreshold} shows how the cost/accuracy metric values for each group, as well as the corresponding difference (discrimination) are changing as more protected group samples are considered for model training. We use 3-fold cross-validation and since we randomly generate $50$ different samples for every size value, the plots are shown with error bars. As expected, the cost/accuracy metric value for male group maintains the same mean while for female group it is changing. Interestingly, according to all cost/accuracy metrics (except AUC), discrimination is increasing as data is more balanced. Figure~\ref{fig:dutchThreshold} shows the results with logistic regression, but the pattern is similar for other classification algorithms (Figure~\ref{fig:dutchThreshold_all} in Appendix~\ref{sec:a_threshold}) and for other benchmark datasets (Figure~\ref{fig:adultThreshold_all} in Appendix~\ref{sec:a_threshold}). This counterintuitive behavior is also observed for the reverse experiment where the protected group (female) sample size is fixed ($100$ samples) while the privileged group (male) is under represented and more samples are collected and considered in the training (Figure~\ref{fig:dutchThreshold_inv}). It is important to mention that in all previous experiments, selecting samples to balance the training set is performed randomly to simulate, as accurately as possible, data collection in real scenarios. 
%{\color{blue}
The fairness enhancing potential of adding more samples for the sensitive group depends on the initial fairness characteristics of the data and the goal of the classifier. Wang et al.~\cite{Wang2018BalancedDA} point out that adding more samples of the minority group to the data increases predictive accuracy and fairness specifically in the classification tasks, where sensitive attribute is part of the output of classification, for example face recognition~\cite{buolamwini2018gender}. %}

\begin{figure*}[t!]
    \centering
    \includegraphics[height=1.6in, width=7.2in]{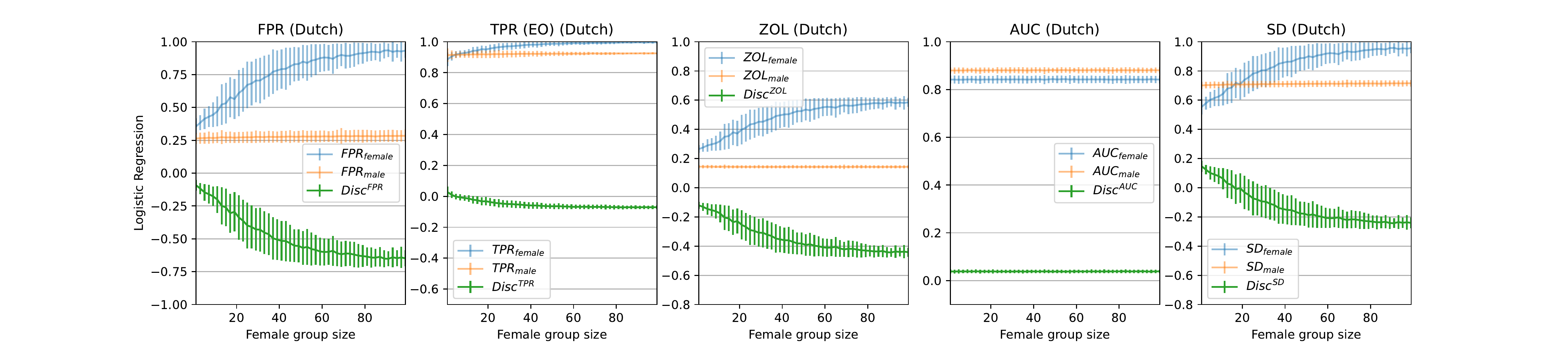}
    \caption{Discrimination while augmenting the training set with only positive outcome female group samples. The male group size is fixed at $100$. Dataset is Dutch Census and training algorithm is logistic regression.}
    \label{fig:dutchThreshold_sel}
\end{figure*}

If, however, training set is balanced by selecting a specific type of samples, in particular, protected group samples with positive outcome, discrimination will be decreasing as data gets balanced (Figure~\ref{fig:dutchThreshold_sel}). In all three experiments (collecting more protected group samples randomly, collecting more unprotected group samples randomly, and collecting only positive outcome protected group samples), the importance of the sensitive feature (Sex) in the prediction (\textit{shap} explanation~\cite{shap}) behaves the same way (Figure~\ref{fig:sensImportance} in Appendix~\ref{sec:a_threshold}), that is, it contributes more to the learned model as the data is more balanced.% \todo[inline]{It asks for a sentence saying how ti behaves in summary, namely that the importance of the sensitive feature increases.}

% \subsection{Sensitive feature importance}

% % \begin{figure*}[t!]
% %     \centering
% %     \begin{subfigure}[t]{0.5\textwidth}
% %         \centering
% %         \includegraphics[height=2in]{figures/SHAP-adult-compas-sizes.pdf}
% %         \caption{Varying sample size}
% %     \end{subfigure}%
% %     ~ 
% %     \begin{subfigure}[t]{0.5\textwidth}
% %         \centering
% %         \includegraphics[height=2in]{figures/SHAP-adult-compas-ratios.pdf}
% %         \caption{Varying sensitive groups ratio}
% %     \end{subfigure}
% %     \caption{Sensitive feature importance (SHAP)}
% %     \label{fig:shap}
% % \end{figure*}

% \subsection{Effect of training set size on discrimination}

\section{Conclusion}

A very common source of bias in machine learning is to use a limited size or imbalanced training set. This paper defines $SSB$ and $URB$ to capture these variants. In the light of empirical analysis on benchmark datasets and using off the shelf classification algorithms, we made three important observations. First, discrimination metrics defined using $AUC$ and $ZOL$ (which consider the trade-off between precision and recall) are more resilient to sampling biases than discrimination defined using $FPR$ and $TPR$ (equal opportunity). Consequently, in presence of limited size or imbalanced training data, it is recommended to use fairness metrics based on the trade-off between precision and recall (e.g. equalized odds~\cite{hardt2016equality}) to reliably estimate discrimination. 
Second, for regression problems, discrimination defined in terms of $MSE$ is significantly affected by variance for extremely small or imbalanced training sets. Hence, it is recommended to treat discrimination values with caution in such cases. Third, collecting more samples of the extremely underrepresented group according to the population distribution will typically amplify discrimination rather than reducing it. However, collecting more data, allows to measure discrimination more reliably.

\section*{Acknowledgments}
This work was supported by the European Research Council (ERC) project HYPATIA under the European Union’s Horizon 2020 research and innovation programme. Grant agreement n. 835294.

%Bibliography
% \bibliographystyle{unsrt}  
% \bibliography{references}  
\bibliographystyle{abbrv}
\bibliography{texz-bibFile}

\newpage
\appendix

\section{Appendix}

% \subsection{Related Work}
% \label{sec:a_related}
% \input{text-03-relatedwork}

% \newpage

\subsection{Additional plots for the magnitude of SSB and URB (Sections~\ref{sec:exp_ssb} and~\ref{sec:exp_urb})}
\label{sec:a_ssb_urb}

\begin{figure}[H]
    \centering
    \includegraphics[height=8in, width=7.6in]{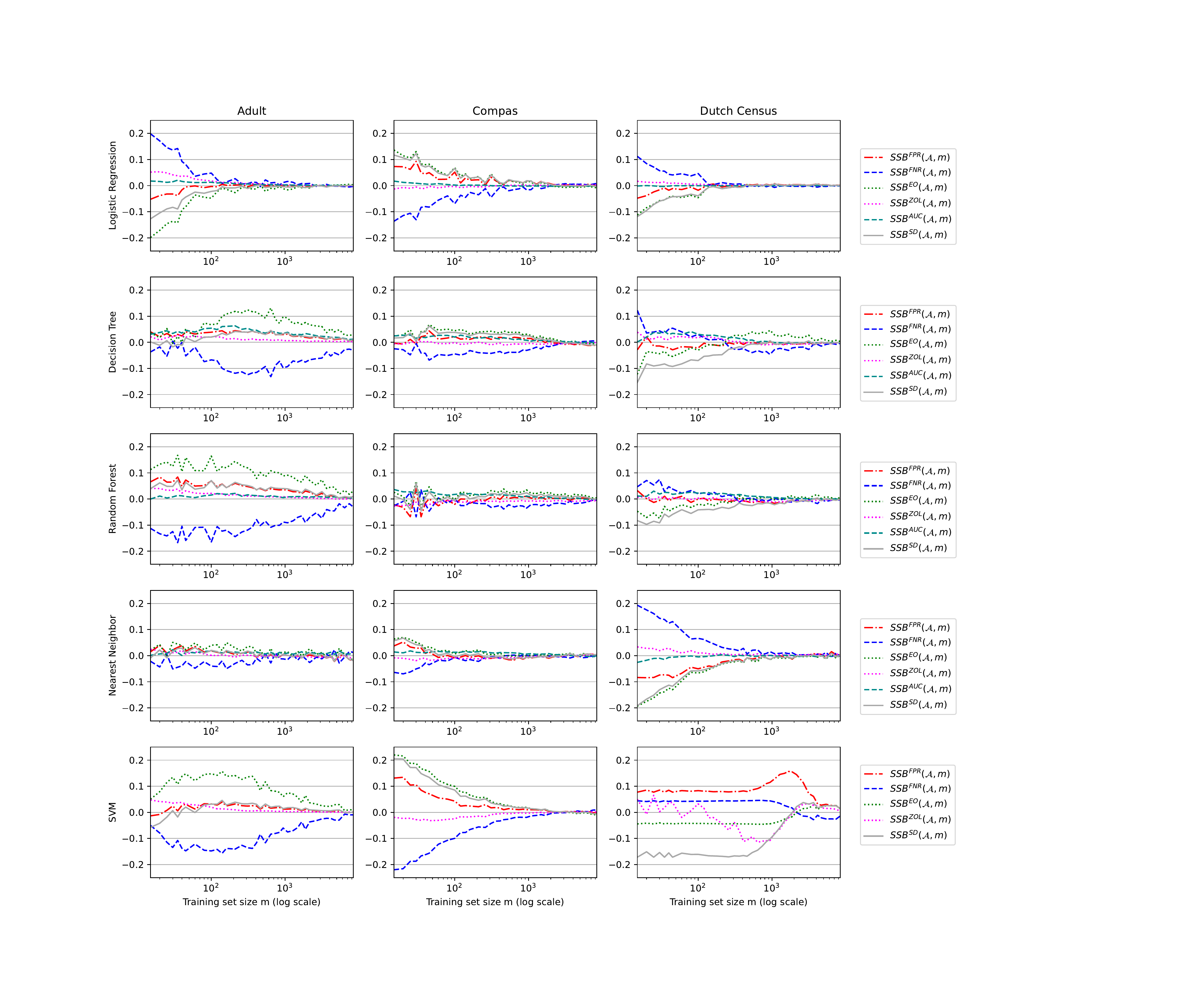}
    \caption{Magnitude of sample size bias (SSB) for increasing size of the training data.}
    \label{fig:all_ssb}
\end{figure}

\begin{figure}[H]
    \centering
    \includegraphics[height=8in, width=7.6in]{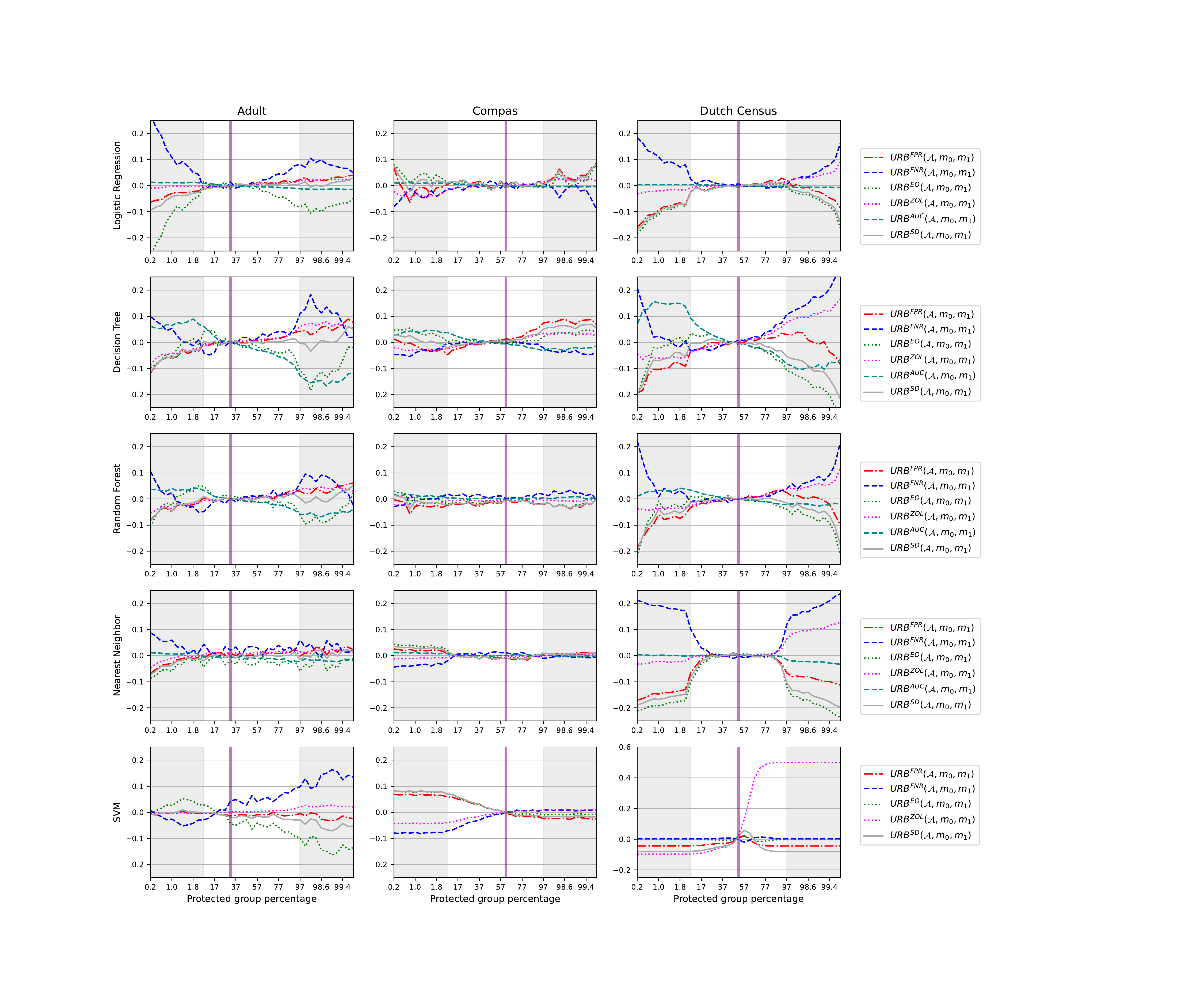}
    \caption{Underrepresentation Bias (URB) for different ratios of sensitive groups. The training set size is fixed ($1000$). The horizontal bar represents the same ratio as the population. The shaded sections indicate a focus on the extreme proportions (less than $2\%$ and more than $98\%$).}
    \label{fig:all_URB}
\end{figure}

\newpage

\subsection{Additional plots for the effect of collecting more samples on discrimination (Section~\ref{sec:exp_threshold})}
\label{sec:a_threshold}

\begin{figure}[H]
    \centering
    \includegraphics[height=8in, width=7.3in]{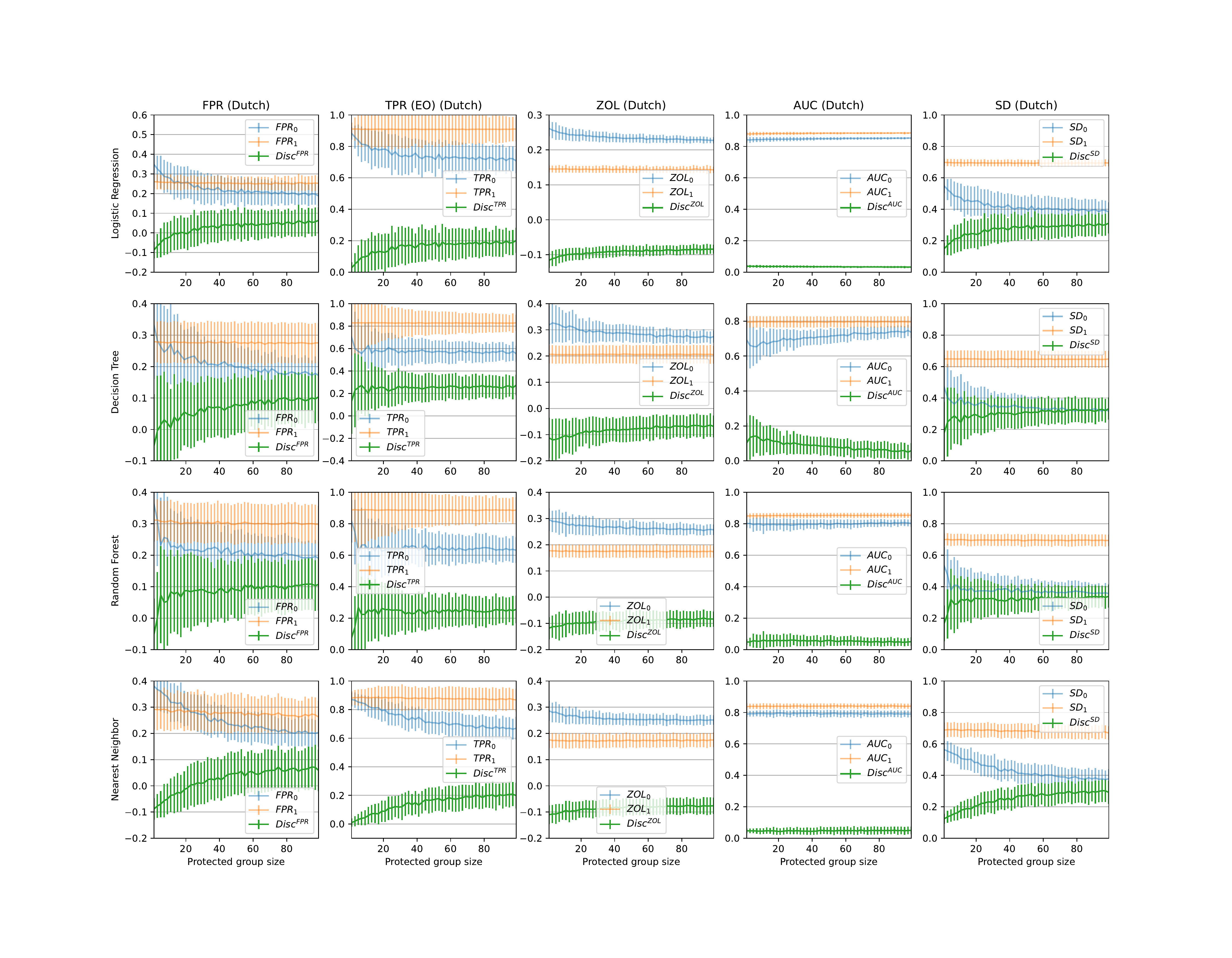}
    \caption{Discrimination values for the Dutch Census dataset while increasing the size of the protected group.}
    \label{fig:dutchThreshold_all}
\end{figure}

\begin{figure}[H]
    \centering
    \includegraphics[height=7.5in, width=7.3in]{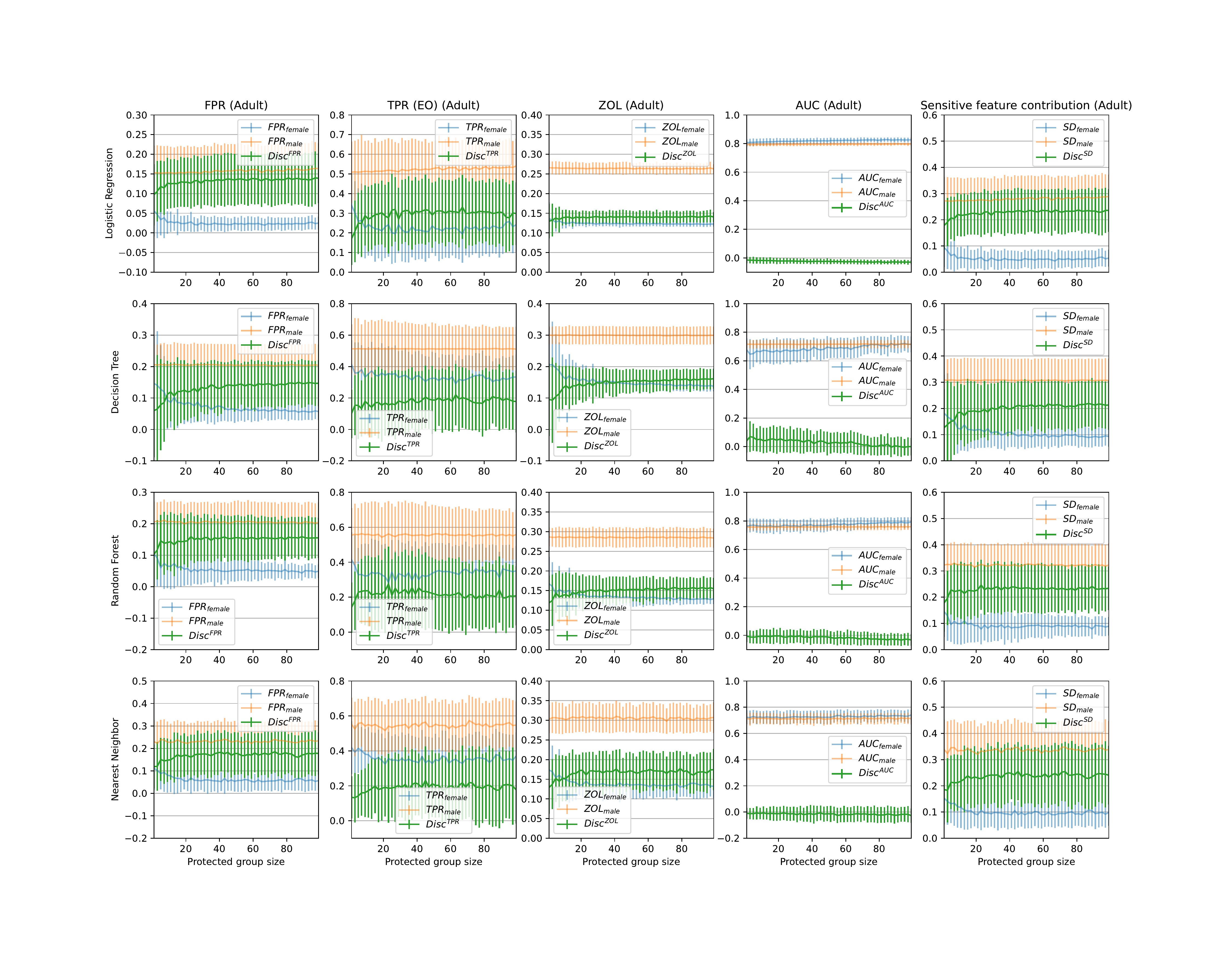}
    \caption{Discrimination value for the Adult dataset while increasing the size of the protected group.}
    \label{fig:adultThreshold_all}
\end{figure}

\begin{figure}[H]
    \centering
    \includegraphics[height=2.3in, width=7in]{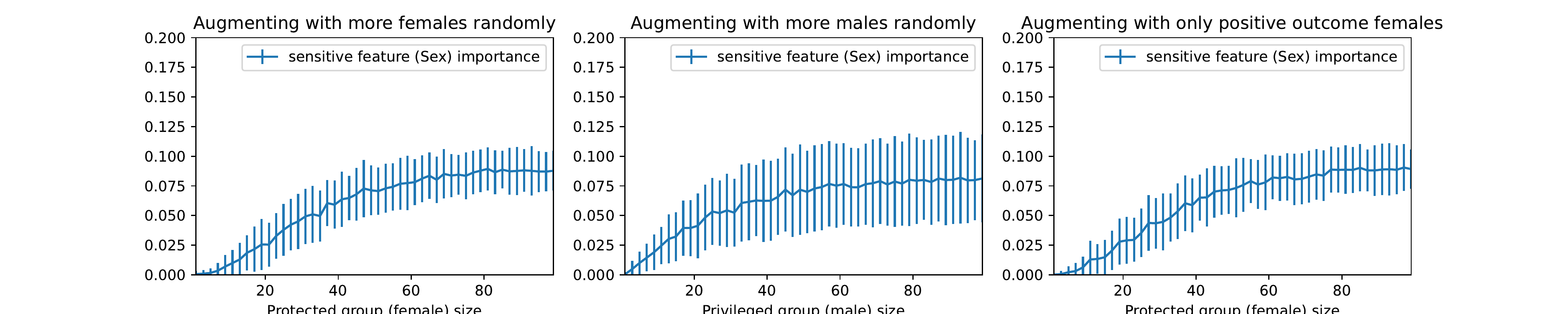}
    \caption{Sensitive feature (Sex) importance observed in the experiments of Section~\ref{sec:exp_threshold}.}
    \label{fig:sensImportance}
\end{figure}

\newpage
\subsection{Decomposing and bounding statistical disparity}
\label{sec:sd}

Statistical disparity is the simplest discrimination metric and it corresponds to the difference in the expected outcomes between groups:
\begin{definition}[Statistical Disparity]
\begin{align}
Disc^{SD}(\Yhs) & = \ep_{\mathcal{X}}[\Yhs | A=a_1] - \ep_{\mathcal{X}}[\Yhs | A=a_0] \nonumber \\
& = \expe{\bx \in G_1}[\hfs(\bx)] - \expe{\bx \in G_0}[\hfs(\bx)] \nonumber 
\end{align}
\end{definition}

$Disc^{SD}(\Yhs)$ is a biased estimation of the \textit{true} value $Disc^{SD}(Y)$. The following theorem states that the error in estimating statistical disparity can be bounded where the bounds are expressed in terms of noise, bias, and variance.

\begin{theorem}
The error in estimating statistical disparity is bounded as follows:
\begin{align}
 |Disc^{SD}(\Yhs) - Disc^{SD}(Y)| & \leq   (\overline{N}^{AL}_{a_1}(\Yhs) - \overline{N}^{AL}_{a_0}(\Yhs)) +  (\overline{B}^{AL}_{a_1}(\Yhs) - \overline{B}^{AL}_{a_0}(\Yhs)) + \nonumber \\
 & (\overline{V}^{AL}_{a_1}(\Yhs) - \overline{V}^{AL}_{a_0}(\Yhs)) \nonumber \\
 |Disc^{SD}(\Yhs) - Disc^{SD}(Y)| & \geq  \max ( \nonumber \\
 & \quad (\overline{N}^{AL}_{a_1}(\Yhs) - \overline{N}^{AL}_{a_0}(\Yhs)) -  (\overline{B}^{AL}_{a_1}(\Yhs) - \overline{B}^{AL}_{a_0}(\Yhs)) - \nonumber \\
 &\quad (\overline{V}^{AL}_{a_1}(\Yhs) - \overline{V}^{AL}_{a_0}(\Yhs)), \nonumber \\
 & \quad  (\overline{B}^{AL}_{a_1}(\Yhs) - \overline{B}^{AL}_{a_0}(\Yhs)) - (\overline{N}^{AL}_{a_1}(\Yhs) - \overline{N}^{AL}_{a_0}(\Yhs))  - \nonumber \\
 & \quad (\overline{V}^{AL}_{a_1}(\Yhs) - \overline{V}^{AL}_{a_0}(\Yhs)), \nonumber \\
 & \quad  (\overline{V}^{AL}_{a_1}(\Yhs) - \overline{V}^{AL}_{a_0}(\Yhs)) - (\overline{B}^{AL}_{a_1}(\Yhs) - \overline{B}^{AL}_{a_0}(\Yhs)) - \nonumber \\
 & \quad (\overline{N}^{AL}_{a_1}(\Yhs) - \overline{N}^{AL}_{a_0}(\Yhs)) ) \nonumber 
\end{align}
where:
\begin{itemize}
   \item[$\circ$] $\overline{N}^{AL}_{a}(\Yhs) = \ep_{\bx \in \mathcal{X}} [N^{AL}(\bx)|A=a]$ 
    \item[$\circ$] $\overline{B}^{AL}_{a}(\Yhs) = \ep_{\bx \in \mathcal{X}} [B^{AL}(\bx)|A=a]$ 
    \item[$\circ$] $\overline{V}^{AL}_{a}(\Yhs) = \ep_{\bx \in \mathcal{X}} [(1-2\times B^{AL}(\bx))\times V^{AL}(\bx)|A=a]$
\end{itemize}
\end{theorem}

\begin{proof} The proof is based on the triangle inequality of metrics. Recall that a metric is a function of two arguments ($dist(x,y)$) that satisfy minimality ($\forall x,y, dist(x,y) \geq dist(x,y)$), symmetry ($\forall x,y, dist(x,y) = dist(y,x)$), and triangle inequality ($\forall x,y,z dist(x,z) + dist(z,x) \geq dist(x,y)$). The full proof is very similar to the proof in~\cite{domingos2000} (Theorem 7). \qed
\end{proof}

\end{document}